\documentclass{article}
\usepackage{spconf,amsmath,graphicx}
\usepackage{algorithmic}
\usepackage{booktabs}
\usepackage{amsmath}
\usepackage{setspace}
\usepackage{array,caption}
\usepackage[labelfont=bf]{caption}
\usepackage{CJKutf8}
\usepackage{amssymb}
\usepackage{tablefootnote}
\usepackage{multirow}
\usepackage{tabularx, booktabs}
\usepackage{tablefootnote}
\usepackage{seqsplit}

\title{Leveraging Biases in Large Language Models: ``bias-kNN'' for Effective Few-Shot Learning}
\name{Yong Zhang\textsuperscript{\rm 1 \textdagger}, Hanzhang Li\textsuperscript{\rm 1 \rm 2 \textdagger} , Zhitao Li\textsuperscript{\rm 1},  Ning Cheng\textsuperscript{\rm 1 *}, Ming Li\textsuperscript{\rm 1 \rm 3}, Jing Xiao\textsuperscript{\rm 1}, Jianzong Wang\textsuperscript{\rm 1}}
\address{
  \textsuperscript{\rm 1}Ping An Technology (Shenzhen) Co., Ltd., China\\
  \textsuperscript{\rm 2}Lanzhou University, China\\
    \textsuperscript{\rm 3} University of Maryland}
\begin{document}

%
\maketitle

\renewcommand{\thefootnote}{}
\footnotetext{\textdagger Equal contribution}
\renewcommand{\thefootnote}{\arabic{footnote}}
\renewcommand{\thefootnote}{}
\footnotetext{ *Corresponding authors: Ning Cheng (chengning211@pingan.com.cn)}
\renewcommand{\thefootnote}{\arabic{footnote}}

\begin{abstract}
Large Language Models (LLMs) have shown significant promise in various applications, including zero-shot and few-shot learning. However, their performance can be hampered by inherent biases. Instead of traditionally sought methods that aim to minimize or correct these biases, this study introduces a novel methodology named ``bias-kNN''. This approach capitalizes on the biased outputs, harnessing them as primary features for kNN and supplementing with gold labels. Our comprehensive evaluations, spanning diverse domain text classification datasets and different GPT-2 model sizes, indicate the adaptability and efficacy of the ``bias-kNN'' method. Remarkably, this approach not only outperforms conventional in-context learning in few-shot scenarios but also demonstrates robustness across a spectrum of samples, templates and verbalizers. This study, therefore, presents a unique perspective on harnessing biases, transforming them into assets for enhanced model performance.

\end{abstract}

\begin{keywords}
LLM, Model Bias, Bias Leverage, kNN Methods, Zero-Shot Learning, Few-shot Learning
\end{keywords}
\vspace{-2.5mm}

\section{Introduction}

Large language models (LLMs) have emerged as powerful tools showcasing impressive zero-shot and few-shot capabilities \cite{petroni2019language, brown2020language}. Leveraging templates and verbalizers \cite{schick-schutze-2021-verbalizer} to align an LLM's output probability distribution with task-specific labels allows the model to address downstream classification tasks in zero-shot or few-shot contexts.

\begin{figure}[!tp]
\centering
\includegraphics[width=0.47\textwidth]{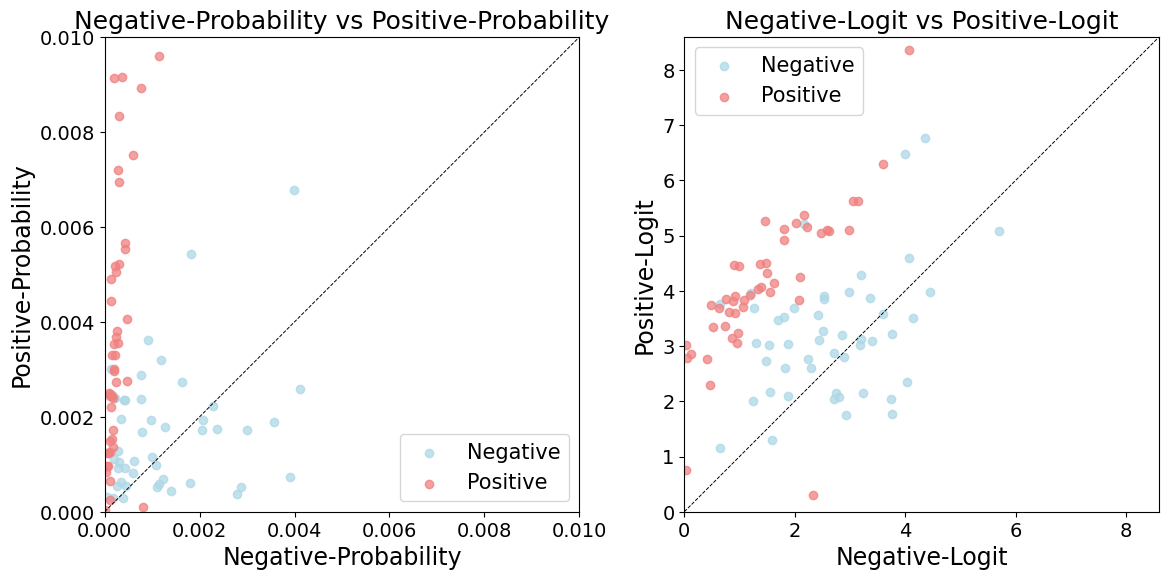}

\caption{Zero-shot probability and logit results from the CR train dataset, visualizing 50 samples each from the Positive and Negative categories. The model exhibits a clear bias towards the Positive category. The dashed line $y=x$ denotes the decision boundary for these categories.}
\label{fig}
\vspace{-6.5mm}

\end{figure}

However, these models are not without their challenges, predominantly stemming from biases. These biases influence both the model's inherent discriminative abilities and its output, skewing probability values for specific categories. Moreover, they can also disrupt conventional decision boundaries, thereby compromising their reliability \cite{zhao2021calibrate, holtzman2021surface}. Research identifies these biases mainly as vanilla label bias, where frequently encountered words during pre-training get prediction preference, and domain label bias, where the bias manifestation varies based on content domain \cite{fei2023mitigatingdomainbias}. Another noteworthy phenomenon is the surface form competition \cite{holtzman2021surface}, where semantically similar words vie for identical probability space, leading to distributional conflicts.







Addressing biases in LLMs necessitates a diverse strategy. Initially, some methods focus on direct bias measurement and recalibration. Take, for instance, Contextual Calibration \cite{zhao2021calibrate} which uses neutral test inputs, such as ``N/A'' to recalibrate model outputs. In a similar vein, Domain-Context Calibration \cite{fei2023mitigatingdomainbias} leverages random in-domain tokens to gauge the bias probability of individual labels. While potent, these approaches sometimes apply a broad-brush correction, occasionally missing the nuanced biases specific to certain test samples.

In another category, methods like PROCA \cite{han2022prototypical} strategize around defining an optimal classification boundary. They draw on the model's contextual insights and employ a Gaussian Mixture Model (GMM) to understand the data spread. Similarly, approaches such as KNN-C \cite{nie2022improvingknn} and kNN-prompting \cite{xu2023knnprompting} harness the model representations, emphasizing its capability for representation over prediction \cite{khandelwal2019knnlm}, to navigate around biases rather than confront them directly.

As depicted in Figure 1, biases in LLMs frequently result in lower probabilities for verbalizers, causing dense clusters and category overlaps. However, the evident directionality differences between categories hint at an intriguing opportunity: utilizing biased outputs together with Nearest Neighbor methods to enhance sample inference.

Moving away from traditional strategies that aim to minimize or correct biases, we present a novel approach termed ``bias-kNN''. In this method, we utilize biased outputs as primary features for kNN, enriched by gold labels. Our evaluations, which encompass various domain text classification datasets and different GPT-2 model sizes \cite{radford2019gpt2}, demonstrate that ``bias-kNN'' not only surpasses the performance of traditional in-context learning in few-shot settings but also exhibits robustness across diverse templates and verbalizers.

The contributions of this paper are:

\begin{itemize}

\vspace{-1.5mm}

\item We unveil a pioneering approach, ``bias-kNN'', which diverges from traditional strategies that aim to minimize or correct biases. Instead, this method capitalizes on biased outputs by using them as primary features for kNN, further enriched by gold labels.

\vspace{-1.5mm}

\item Our rigorous evaluations cover a range of domain text classification datasets and span different GPT-2 model sizes, reinforcing the versatility and adaptability of the ``bias-kNN'' approach.
\vspace{-1.5mm}

\item The ``bias-kNN'' approach consistently outperforms traditional in-context learning in few-shot scenarios and exhibits enhanced stability across varied labeled data samples. Furthermore, its proven robustness with diverse templates and verbalizers highlights its resilience and broad applicability.

\end{itemize}

    \vspace{-3mm}

\section{Methodology}

\vspace{-2.5mm}

In this section, we detail our kNN modeling technique that enhances text classification by harnessing biases in logit outputs from language models. The method's structure is depicted in Figure 2.
    \vspace{-1mm}

\subsection{Bias Output based kNN Modeling}
\vspace{-1.5mm}

Our approach is based on the idea that model outputs, even with their challenges, have valuable distinguishing features. With this in mind, we adapted the kNN method.

Given a model \( \mathcal{M}_{\theta} \), a domain-specific labeled dataset \( \mathcal{A}=\left\{\left(x_i, y_i\right)\right\}_{i=1}^{|A|} \), a template \( \mathcal{T} \), the label set of the domain data \( \mathcal{Y} = \left\{y_j^*\right\}_{j=1}^{|\mathcal{Y}|} \) and a verbalizer $\mathcal{V}$ to map each label word of $\mathcal{Y}$ to a word $v$ in the \( \mathcal{M}_{\theta} \)'s vocabulary. We employed the template \( \mathcal{T} \) to structure each \( x_i \) as \( \mathcal{T}(x_i) \) and feed to the model \( \mathcal{M}_{\theta} \) to get the probability output $\mathcal{P}$, representing the probability $p_{\left(y_j^*|\mathcal{T}(x_i)\right)}$ of each $y_j^*$ in the target label set
$\mathcal{Y}$.

\vspace{-2.5mm}

\begin{equation}
\left\{\textit{p}_{\left(y_j^*| \mathcal{T}(x_i)\right)}\right\}_{j=1}^{|\mathcal{Y}|} = \mathrm{M}_\theta(\mathrm{y} \mid \mathbf{x}) \propto \mathrm{M}_{\theta}(\mathrm{V}(\mathrm{y}) \mid \mathcal{T}(x))
\end{equation}

 \begin{figure}[!tp]
\centering
\includegraphics[width=0.48\textwidth]{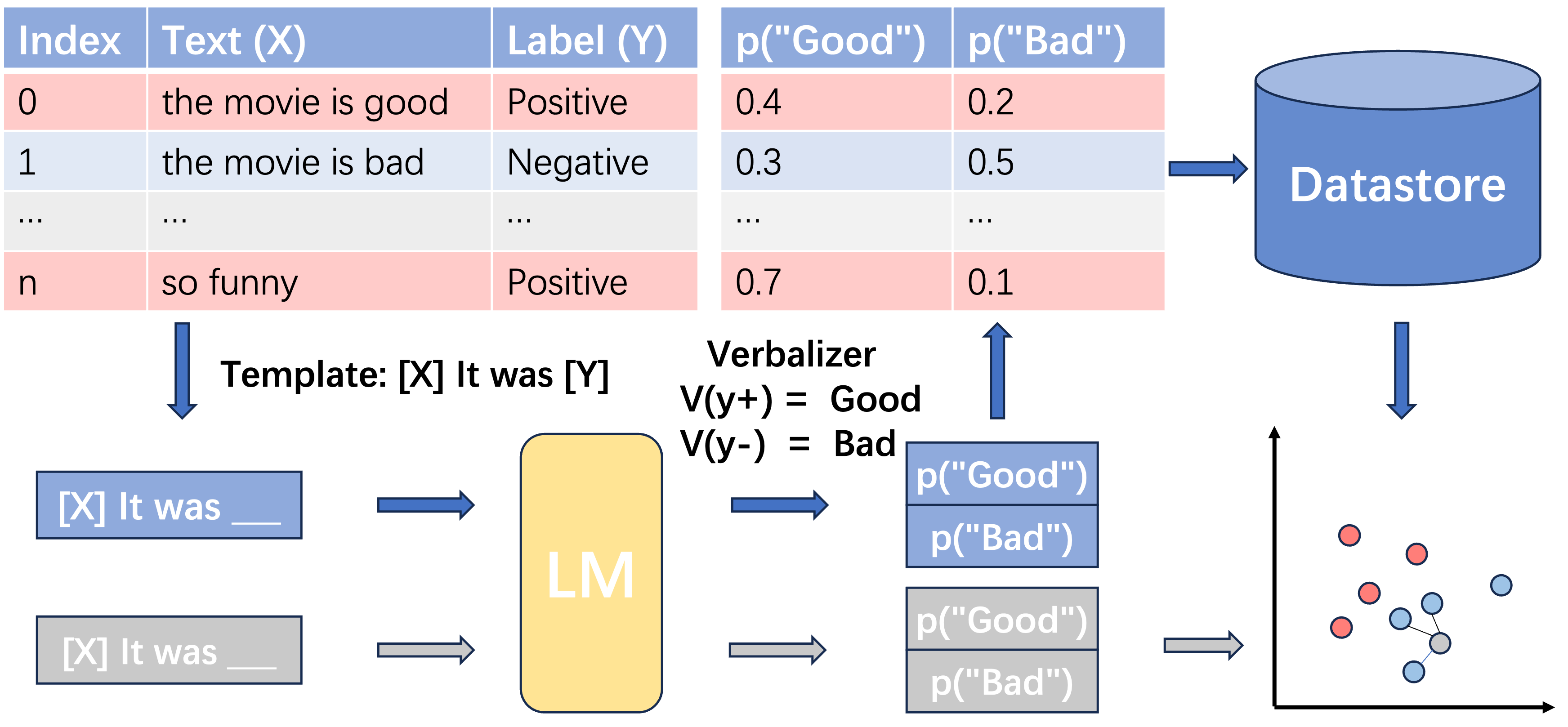}
\caption{The architecture of our proposed model}
\vspace{-5mm}

\end{figure}

Using the model's output probabilities, which encapsulate the biases, we transform them into features that can be utilized in a kNN framework. During the prediction phase, our approach retrives the k most similar samples from the datastore $\mathcal{K}$ =  $\mathrm{kNN}\left(\mathcal{A}, \mathcal{P}\right)$ using the cosine distance metric. The definitive label $y_{\text {pred }}$ for the input sample is then ascertained through a majority vote.

\begin{equation}
y_{\text {pred }}=\underset{y_j^* \in \mathcal{Y}}{\arg \max } \sum_{i \in \text{NN}^k\left(\mathcal{K},x_i\right)} \mathbf{1}\left(y_i=y_j^*\right)
\end{equation}

In essence, our method harnesses the biases typically found in LLM outputs, transforming potential shortcomings into features that empower a kNN-based classification approach.

\vspace{-3mm}

\section{Experiment}
\vspace{-2mm}

\subsection{Setup}

\subsubsection{Datasets}

We evaluated our approach on six classification tasks, spanning four distinct task families and covering a wide range of data domains. For \textbf{Sentiment Classification}, we employed datasets like the Stanford Sentiment Treebank (SST-2) \cite{socher2013recursive_sst2}, Movie Reviews (MR) \cite{pang2005seeing-mr}, and CommitmentBank (CR) \cite{hu2004mining-cr}, all of which categorize sentiments into binary classes. \textbf{Topic Classification} was addressed using the AGNews dataset \cite{zhang2015character_agnews}, which classifies articles into one of four news categories. The \textbf{Subjectivity Classification} was conducted using the Subj dataset \cite{pang2004sentimental-subj}, differentiating sentences from movie statements into subjective or objective categories. Lastly, for \textbf{Entailment Analysis}, we harnessed the Recognizing Textual Entailment (RTE) dataset \cite{dagan2005pascal-rte}, a resource specifically curated for textual entailment tasks.

\vspace{-3.5mm}

\subsubsection{Evaluation}
    \vspace{-1mm}

Our evaluation was designed to rigorously and comprehensively assess the effectiveness of our proposed method over the diverse datasets. For each dataset, we present results detailing the mean, minimum, and standard deviation of accuracy.

\vspace{-3mm}

\begin{table*}[!htbp]
\centering
\vspace{-1mm}

\caption{Main results on classification tasks}
\footnotesize
\vspace{-1mm}

\begin{tabularx}{\textwidth}{llXXXXXX}
\toprule
\multicolumn{2}{c}{\textbf{Method}} & \textbf{SST-2} & \textbf{MR} & \textbf{CR} & \textbf{Subj} & \textbf{RTE} & \textbf{AGNews} \\

\midrule

\multirow{3}{*}{GPT-2-medium} 
& Zero-LM & $58.4/58.4/0.0$ & $57.4/57.4/0.0$ & $66.7/66.7/0.0$ & $73.4/73.4/0.0$ & $56.7/56.7/0.0$ & $44.5/44.5/0.0$ \\

& Raw-ICL$_{m = 3}$ & $65.9/50.9/13.4$ & $66.2/50.8/9.8$ & $71.4/38.1/13.4$ & $46.6/38.8/3.0$ & $48.5/45.8/1.9$ & $54.8/43.5/5.3$ \\
& bias-kNN$_{m = 3}$ & $75.3/71.6/2.4$ & $74.0/69.0/3.1$ & $77.7/73.4/3.0$ & $72.0/67.4/2.3$ & $48.1/44.0/3.3$ & $52.2/50.1/2.0$ \\
\midrule
\multirow{3}{*}{GPT-2-large} 
& Zero-LM & $75.0/75.0/0.0$ & $71.1/71.1/0.0$ & $68.5/68.5/0.0$ & $55.2/55.2/0.0$ & $53.4/53.4/0.0$ & $63.4/63.4/0.0$ \\
& Raw-ICL$_{m = 3}$ & $63.3/50.9/14.7$ & $62.3/50.0/12.1$ & $69.9/62.0/9.6$ & $57.1/48.5/7.8$ & $54.9/51.3/2.0$ & $63.1/42.1/10.0$ \\
& bias-kNN$_{m = 3}$ & $79.5/77.8/2.7$ & $76.8/75.0/1.7$ & $83.9/81.0/2.1$& $51.8/45.0/6.0$ & $48.9/43.7/2.4$ & $67.3/66.0/0.9$ \\

\midrule
\multirow{3}{*}{GPT-2-XL} 
& Zero-LM & $67.2/67.2/0.0$ & $65.0/65.0/0.0$ & $66.0/66.0/0.0$ & $58.6/58.6/0.0$ & $53.4/53.4/0.0$ & $56.1/56.1/0.0$ \\

& Raw-ICL$_{m = 3}$ & $61.9/51.0/10.1$ & $55.7/50.0/9.1$ & $68.7/62.2/6.1$ & $30.8/21.9/5.9$ & $53.2/52.7/1.1$ & $78.5/71.3/3.2$ \\
& bias-kNN$_{m = 3}$ & $75.9/66.2/5.7$ & $79.2/74.5/2.3$ & $81.7/72.9/5.0$ & $57.6/36.9/9.7$ & $50.6/45.8/3.5$ & $56.6/53.6/1.9$ \\\bottomrule
\end{tabularx}

\footnotesize{ Notes: The three digits in each cell represent the mean, min, and standard deviation of accuracy.}
\vspace{-4.5mm}

\end{table*}

\subsubsection{Baselines}
    \vspace{-1mm}

To ascertain the efficacy of our methodology, we juxtaposed it against several benchmark techniques:

\begin{itemize}

    \item \textbf{Zero-LM}: By leveraging manual prompts and verbalizers, this method gleans predictions directly from the language model (LM) without necessitating training samples. The chosen label is the one with the highest outcome.
    
    \vspace{-3mm}

    \item \textbf{Raw-ICL}: This is a straightforward implementation of In-Context Learning (ICL). It employs \( m \) demonstration samples in conjunction with a prompt and verbalizer. The label is pegged to the most pronounced outcome. Notably, for comparative purposes, the same labeled data sample used for bias-kNN serves as a demonstration.


    \vspace{-3mm}

\end{itemize}

\vspace{-3.5mm}

\subsubsection{Implementation Details}

We utilized three GPT-2 variants distinguished by their capacities: GPT-2-medium(0.3B), GPT-2-large(0.8B), and GPT-2-XL(1.5B). For our kNN modeling, the number of training samples for each category is denoted as $\textbf{\textit{m}}$. Samples of sizes 2, 3, 4, 5, 6, 7, 8, 16, 32, and 64 were selected to construct our kNN datastore. Both construction and inference utilized the same model. The chosen templates and verbalizers are itemized in Table 1. To offset the potential influence of randomness, we executed five random samplings for kNN datastore construction for each \( m \) and a separate set of five for the Raw-ICL demonstration order given the specified \( m \). For a fair assessment, identical templates and verbalizers were adopted for Zero-LM and Raw-ICL. The kNN-Prompt results were sourced directly from the relevant paper. In all cases, the number of neighbors, k, was set to 3.

\vspace{-3mm}

\begin{table}[!htbp]
\setlength{\tabcolsep}{0.5pt}  
\caption{Templates and verbalizers}

\scalebox{0.82}{
\begin{tabular}{lll}
\hline 
\textbf{Dataset} & \textbf{Template} & \textbf{Verbalizer} \\
\hline 
SST2 & \multirow{3}{*}{$\begin{array}{l}\text { Review: [X] Sentiment: [Y] }\end{array}$} & \multirow{3}{*}{Positive, Negative} \\
MR & & \\
CR & & \\
\hline 
AGNews & $\begin{array}{l}\text { Input: [X] Type: [Y] }\end{array}$ & \parbox[t]{3cm}{World, Sports,\\ Business, Tech} \\
\hline 
Subj & $\begin{array}{l}\text { Input: [X] Type: [Y] }\end{array}$ & Objective, Subjective \\
\hline 
RTE & $\begin{array}{l}\text { [X1] Hypothesis: [X2] Prediction: [Y] }\end{array}$ & True, False \\
\hline
\end{tabular}}

\vspace{-6mm}

\end{table}

\subsection{Main Results}
    \vspace{-1mm}

Table~1 presents a detailed comparison of our method based on the GPT-2 models against current baselines over various datasets. The performance of bias-kNN improves with an increase in the value of \( m \), displaying higher mean accuracy and smaller standard deviation, as illustrated in the subsequent figures.

In the table, we report the smallest \( m \) that yields good performance. For the GPT-2-large model, it is evident that with \( m=3 \), the minimum accuracy values of bias-kNN consistently surpass those of Zero-LM and Raw-ICL across all classification datasets. Moreover, the bias-kNN method demonstrates greater stability compared to the Raw-ICL method. This suggests that utilizing labeled data via bias-kNN might be preferable to ICL in certain scenarios. It is worth noting that both Raw-ICL and bias-kNN encounter challenges with the Subj and RTE tasks, corroborating findings from a previous study \cite{xu2023knnprompting}.

Different sizes of the gpt models exhibit patterns analogous to the GPT-2-large model, albeit with minor variations in accuracy across datasets. Interestingly, we observe that, in some cases, smaller models outperform their larger counterparts, which could be due to the randomness of instance selection and the effects of the template and verbalizer, as shown in  \cite{zhao2021calibrate, holtzman2021surface}.

\vspace{-4mm}

\subsection{Ablation Study and Analysis}
    \vspace{-1mm}

In this section, we conduct experiments to analyze the impact of various choices related to templates, verbalizers, output features, and distance metrics of the kNN. We base these experiments on the CR dataset using GPT-2-large and consine distance metric.
\vspace{-3mm}

\subsubsection{Robustness of Templates and Verbalizers}
    \vspace{-1mm}

Selecting an optimal verbalizer, as corroborated by previous studies, can be a demanding task in terms of resources \cite{schick-schutze-2021-verbalizer, schick2021s-verbalizer2, hu2022knowledgeable_verbalizer}. Figure 3 showcases the chosen verbalizers which are prominent token pairs from the vocabulary. Results strongly suggest that bias-kNN outdoes Zero-LM when \( m > 2 \). Remarkably, bias-kNN also outperforms Zero-LM for certain verbalizer choices associated with causal tokens at \( m = 2 \). This highlights the potential of bias-kNN in simplifying the task of selecting an effective verbalizer.

\begin{figure}[!htbp]
\centering

\vspace{-3mm}

\includegraphics[width=0.45\textwidth]{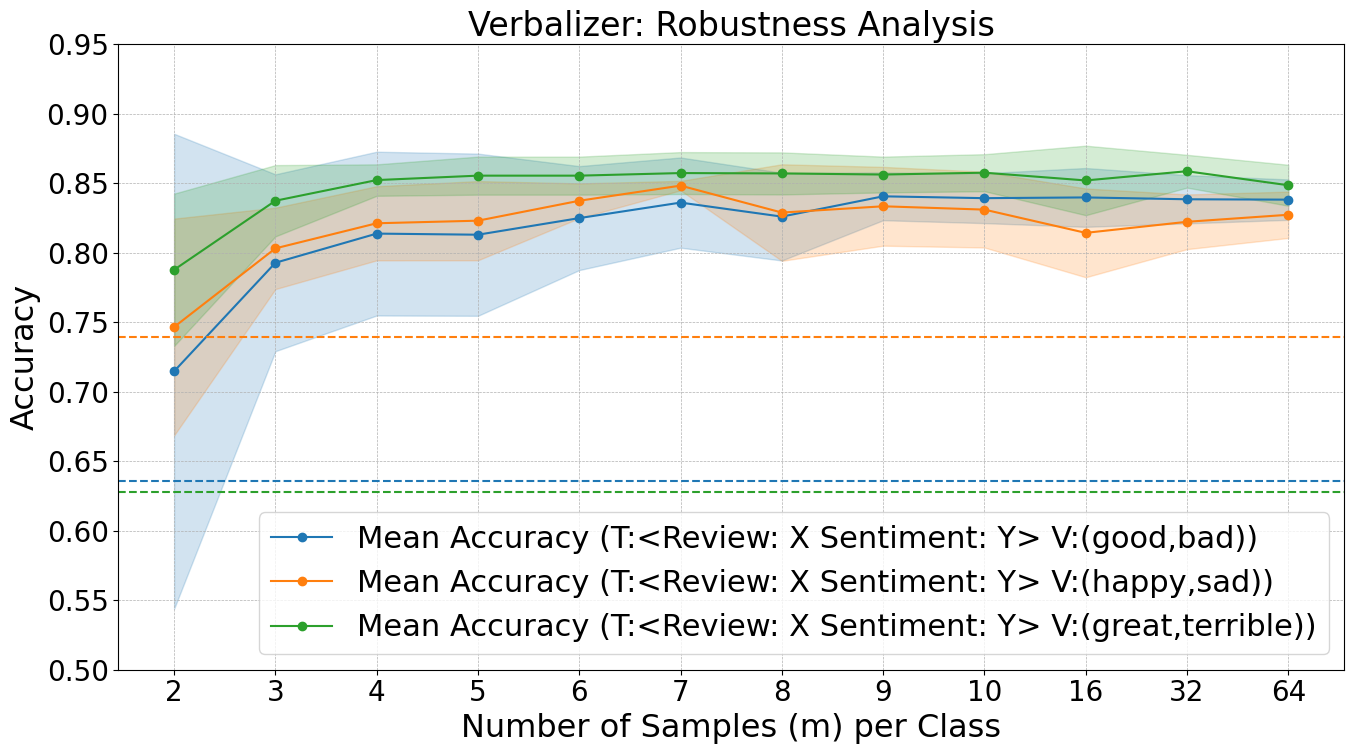}

\vspace{-1mm}

\caption{Verbalizer: robustness analysis. The shaded region denotes the standard deviation. All figures are consistent. Dashed lines of the same color indicate the Zero-LM accuracy for the corresponding settings. }
\label{fig:verbalizer_robustness}
\vspace{-5.5mm}

\end{figure}

Moreover, as depicted in Figure 4, altering templates can lead to variability in the performance of bias-kNN. However, there's a notable enhancement in performance when \( m > 2 \). This further suggests that the bias-kNN approach can alleviate the challenges associated with template selection \cite{shin2020autoprompt_template}. 

\vspace{-2mm}

\begin{figure}[!hbtp]
\centering
\includegraphics[width=0.45\textwidth]{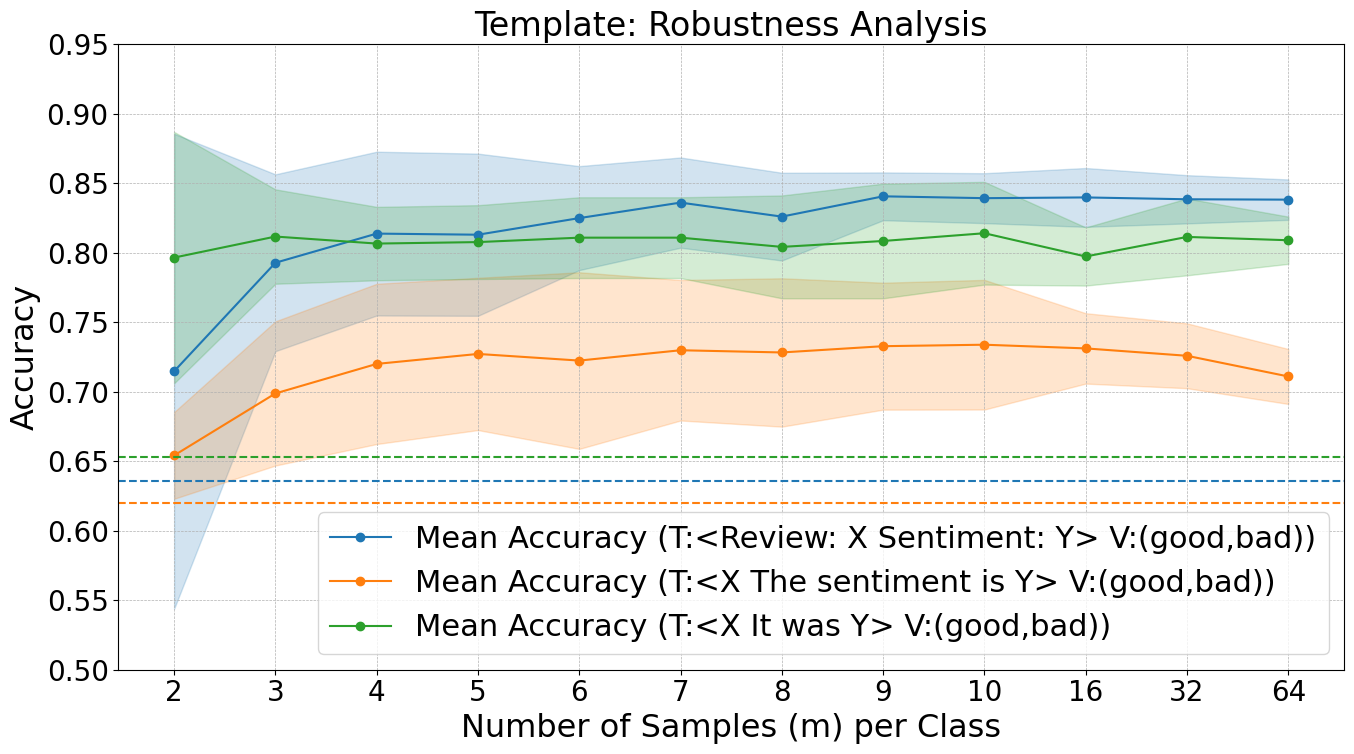}

\vspace{-1mm}

\caption{Template: robustness analysis}
\label{fig:template_robustness}

\vspace{-7.5mm}

\end{figure}

\subsubsection{Biased Logit as a Feature}
Figure 1 illustrates the directionality of the logit. While it aligns more closely with probability, it exhibits a wider distribution. The biased logit reveals a diminished accuracy and an increased standard deviation across various \( m \) values compared to probabilities under the cosine metric.
\vspace{-2mm}

\begin{figure}[!hbtp]
\centering
\includegraphics[width=0.45\textwidth]{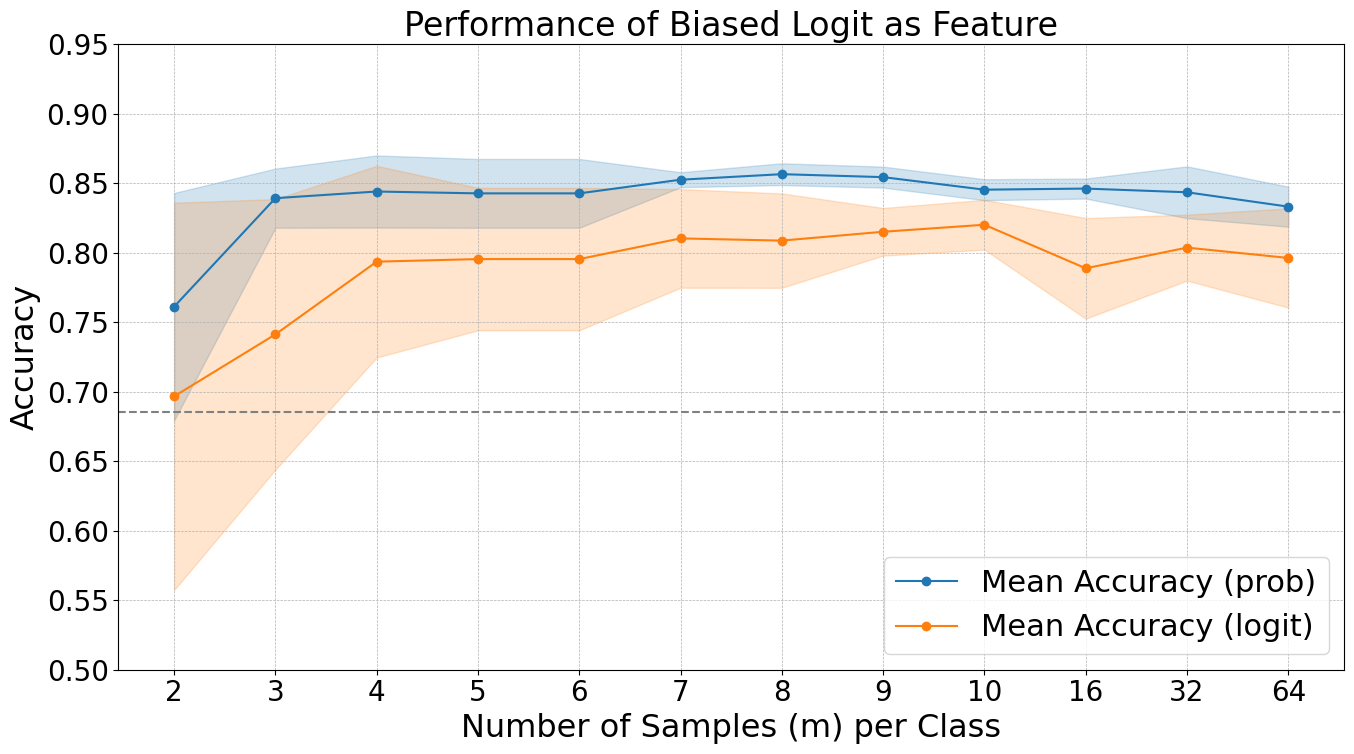}

\vspace{-1mm}

\caption{Performance of biased logit as feature}
\label{fig:biased_logit_performance}

\vspace{-7.5mm}

\end{figure}

\subsubsection{Impact of Distance Metrics}
We evaluated multiple distance metrics including ``euclidean'', ``manhattan'', ``chebyshev'', and ``cosine''. Among these, the ``cosine'' metric showcased the best performance across various setups. This might be attributed to the model's inherent bias that amplifies the likelihood of labels in line with the bias direction.

\vspace{-1.5mm}

\begin{figure}[!hbtp]
\centering
\includegraphics[width=0.45\textwidth]{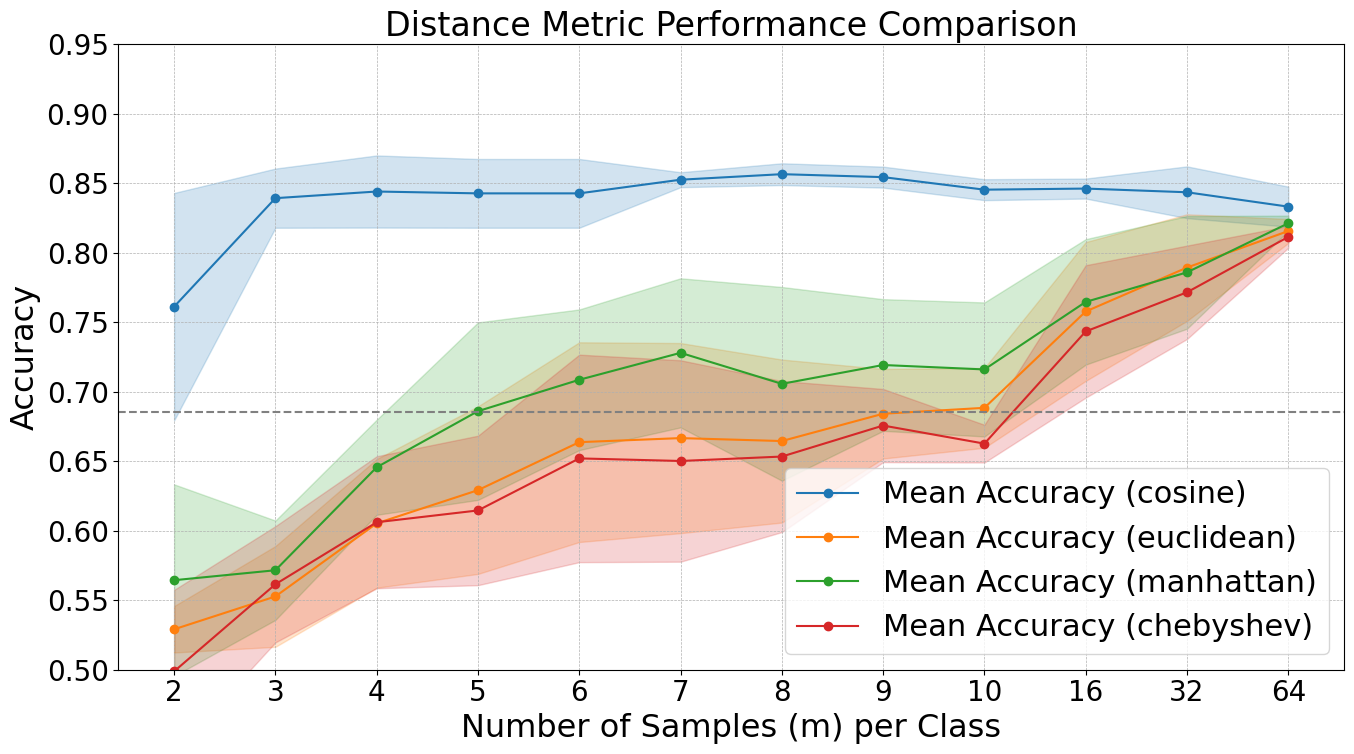}

\vspace{-1mm}

\caption{Distance metric performance comparison}
\label{fig:distance_metric_performance}

\vspace{-7.5mm}

\end{figure}

\vspace{-1.5mm}

\subsubsection{Impact of the Number of Nearest Neighbors (k)}

It becomes evident that increasing the value of $k$ generally does not yield improved performance when $m$ is small. However, it demonstrates superior results when $m$ exceeds 16.
\vspace{-1.5mm}

\begin{figure}[!hbtp]
\centering
\includegraphics[width=0.45\textwidth]{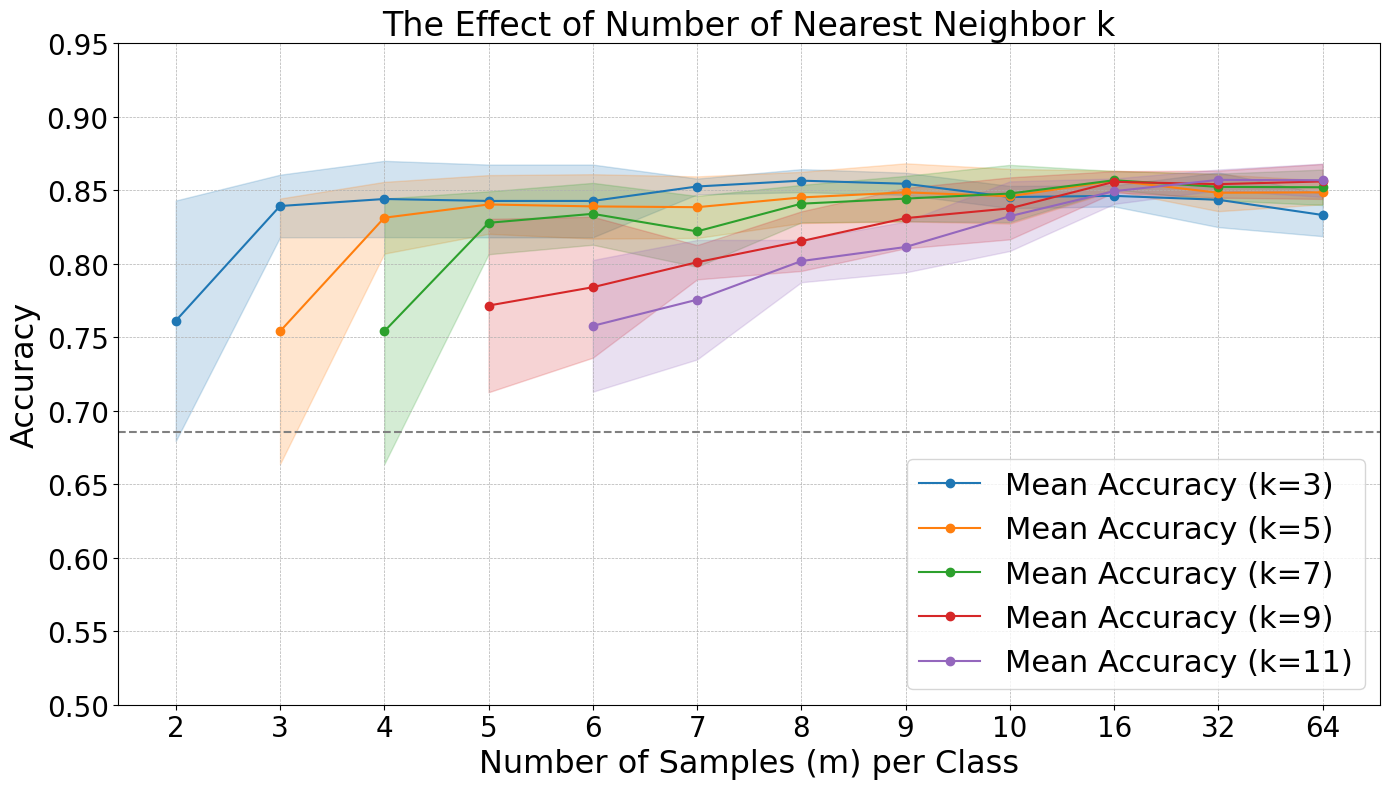}

\vspace{-2mm}

\caption{The effect of number of nearest neighbor k}
\label{fig:The Effect of Number of Nearest Neighbor k}

\vspace{-6.5mm}

\end{figure}

\section{Conclusion}

In this study, we presented the innovative ``bias-kNN'' approach, harnessing biases in large language models to bolster text classification. Through rigorous evaluations across diverse datasets and GPT-2 model variants, our method consistently outperformed conventional in-context learning strategies. The adaptability of ``bias-kNN'' was further underscored by its robust performance over a range of templates and verbalizers. Contrary to the prevailing perception of biases as solely detrimental in machine learning, our research highlights the potential advantages of strategically leveraging them in specific contexts.

\vspace{-2.2mm}
\section{Acknowledgement}
\vspace{-1mm}
This paper is supported by the Key Research and Development Program of Guangdong Province under grant \seqsplit{No.2021B0101400003}. The corresponding author is Ning Cheng from Ping An Technology (Shenzhen) Co., Ltd (chengning211@pingan.com.cn).

\bibliographystyle{IEEEtran}

\bibliography{main} 

\end{document}